\documentclass[]{bytedance}
\usepackage[toc,page,header]{appendix}

\usepackage{newunicodechar}
\newunicodechar{’}{\textquoteright}

\usepackage{amsmath,amsfonts,bm}

\def\eqref#1{equation~\ref{#1}}

\def\1{\bm{1}}

\DeclareMathAlphabet{\mathsfit}{\encodingdefault}{\sfdefault}{m}{sl}
\SetMathAlphabet{\mathsfit}{bold}{\encodingdefault}{\sfdefault}{bx}{n}

\usepackage{url}
\usepackage{array,siunitx,colortbl,tabularx}
\usepackage{wrapfig}
\usepackage{amssymb}

\definecolor{OursRow}{RGB}{232,241,250}
\definecolor{DeltaGreen}{RGB}{31,128,72}
\newcommand{\dpos}{\color{DeltaGreen}}
\newcommand{\grp}[2]{\multicolumn{#1}{@{}l}{\textit{#2}}}
\newcommand{\sz}[1]{{\fontsize{7.5}{9}\selectfont\textcolor{black!60}{#1}}}
\newcommand{\hd}[1]{\fontsize{7.5}{9}\selectfont #1}
\newcommand{\lowhd}[1]{\raisebox{-1.2ex}[0pt][0pt]{#1}}

\definecolor{goodblue}{HTML}{0071bc}
\hypersetup{colorlinks=true,breaklinks,linkcolor=seedblue,citecolor=goodblue,urlcolor=seedblue}

\title{
Open-OmniCoach: Evidence-Enhanced\\[-0.05em]
Trajectory Learning for Omni Video Agents
}

\affiliation{ByteDance OmniCoach Team}
\contribution{Full author list in Contributions}

\abstract{
Training capable omni video agents requires more than diverse question-answer pairs: agents must learn to \textit{reliably find, connect, and correct task-relevant evidence in omni-setting}.
However, we argue that existing data pipelines provide limited control over task-relevant evidence and weak supervision for these capabilities.
In this work, we present \textbf{Open OmniCoach}, a new framework for evidence-seeking omni video agents.
It contains a high-quality dataset, \textbf{OmniCoach-240K}, and a new agentic reinforcement learning (RL) method, {Step-Level Evidence-Guided Policy Optimization (\textbf{SEPO})}.
For the former, OmniCoach-240K comprises 248.2K interaction trajectories, including 230.7K multi-span trajectories and 17.4K reflection trajectories.
We design a three-stage data pipeline by leveraging both fine-grained omni-evidence and the agent's own errors.
Meanwhile, compared with other RL baselines, SEPO provides turn-level credit based on improvements in evidence coverage and temporal localization.
Applying both trajectory data and the SEPO method, we obtain \textbf{Open OmniAgent} models.
The resulting models outperform the baselines across multiple video and audio-visual understanding tasks by significant margins, demonstrating the effectiveness of the Open OmniCoach framework.
In particular, Open OmniAgent-30B improves average accuracy over Qwen3-Omni by 6.3 percentage points across nine benchmarks, with gains of 12.2 and 13.3 points on LVOmniBench and LVBench, respectively.

}

\checkdata[Project Page]{\url{https://open-omnicoach.github.io}}
\checkdata[Code]{\url{https://github.com/Open-OmniCoach/omnicoach}}
\checkdata[Models \& Data]{\url{https://huggingface.co/Open-OmniCoach}}

\begin{document}

\maketitle

\section{Introduction}
\label{sec:intro}

Recently, agent-based approaches have been widely applied to multimodal reasoning problems~\citep{zheng2025deepeyes,wu2025mmsearchr1,hong2025deepeyesv2,yang2025longvt,xing2026omniagent,wang2026vgr,wang2026traceable}, enabling active evidence acquisition and improved comprehension.
In this field, high-quality task data and effective post-training methods are crucial for effective agent training~\citep{tao2025omniagent,xu2026aopagent,xing2026omniagent}.
The former always contains interaction trajectories, while the latter has more optimization potential for better results.

Recent works extend agent-based approaches to video-audio (omni) understanding, with initial efforts constructing tool-calling interaction trajectories~\citep{tao2025omniagent,xu2026aopagent,xing2026omniagent}.
However, these works still provide fewer details on dataset construction and offer less fine-grained evidence control for building high-quality omni-trajectories.
More specifically, how can we make trajectory synthesis more controllable and fine-grained, and enable agents to learn the capabilities needed in omni settings, such as active evidence acquisition and audio-visual evidence binding?
In this way, we can mitigate hallucinations in intermediate interaction steps.

To meet these requirements, we introduce \textbf{Open-OmniCoach}, a new framework for omni video agent training that includes a high-quality training dataset (\textbf{OmniCoach-240K}) and a new RL method (Step-Level Evidence-Guided Policy Optimization, \textbf{SEPO}).
It has \textit{three new key features}:

\begin{enumerate}
    \item \textbf{Generating Trajectory From Fine-Grained Omni-Evidence.}
    Existing omni-modal datasets~\citep{cai2026omnivideo100k,geng2024longvale} are mostly built on videos of a few minutes, rarely requiring an agent to relate distant moments or to locate brief events in long videos.
    Instead, we first compose evidence with task-specific skills across a hierarchy of global events and local audio-visual details, enabling question-answering and temporal grounding tasks that span diverse reasoning types and temporal scales.
    Then, we couple dynamic evidence cropping with multi-span joint interpretation in a trajectory-generation harness, enabling frontier models to crop multiple audio-visual spans in a single tool call and jointly inspect them as reasoning unfolds.

    \item \textbf{Generating Trajectory from the Agent's Own Errors.}
    Existing pipelines supervise agents only with successful trajectories~\citep{xing2026omniagent,zeng2026videoo3,yang2025longvt}, which never show how to recover from the agent's own mistakes, although trained agents still make claims unsupported by the evidence they inspect~\citep{qiu2026videoseal}.
    We, instead, diagnose failed rollouts of our SFT agent at key steps against the inspected audio and frames, and let a teacher resume from the first error with an explicit correction.
    \item \textbf{Crediting Rewards For Each Step.} When a single trajectory-level advantage is applied to every evidence-seeking step, a tool call that uncovers new evidence receives the same advantage as one that repeats an earlier observation.
    Instead, we present an improved version of the RL post-training method, guided by omni-evidence.

\end{enumerate}

\begin{figure}[t]
\centering
\includegraphics[width=\linewidth]{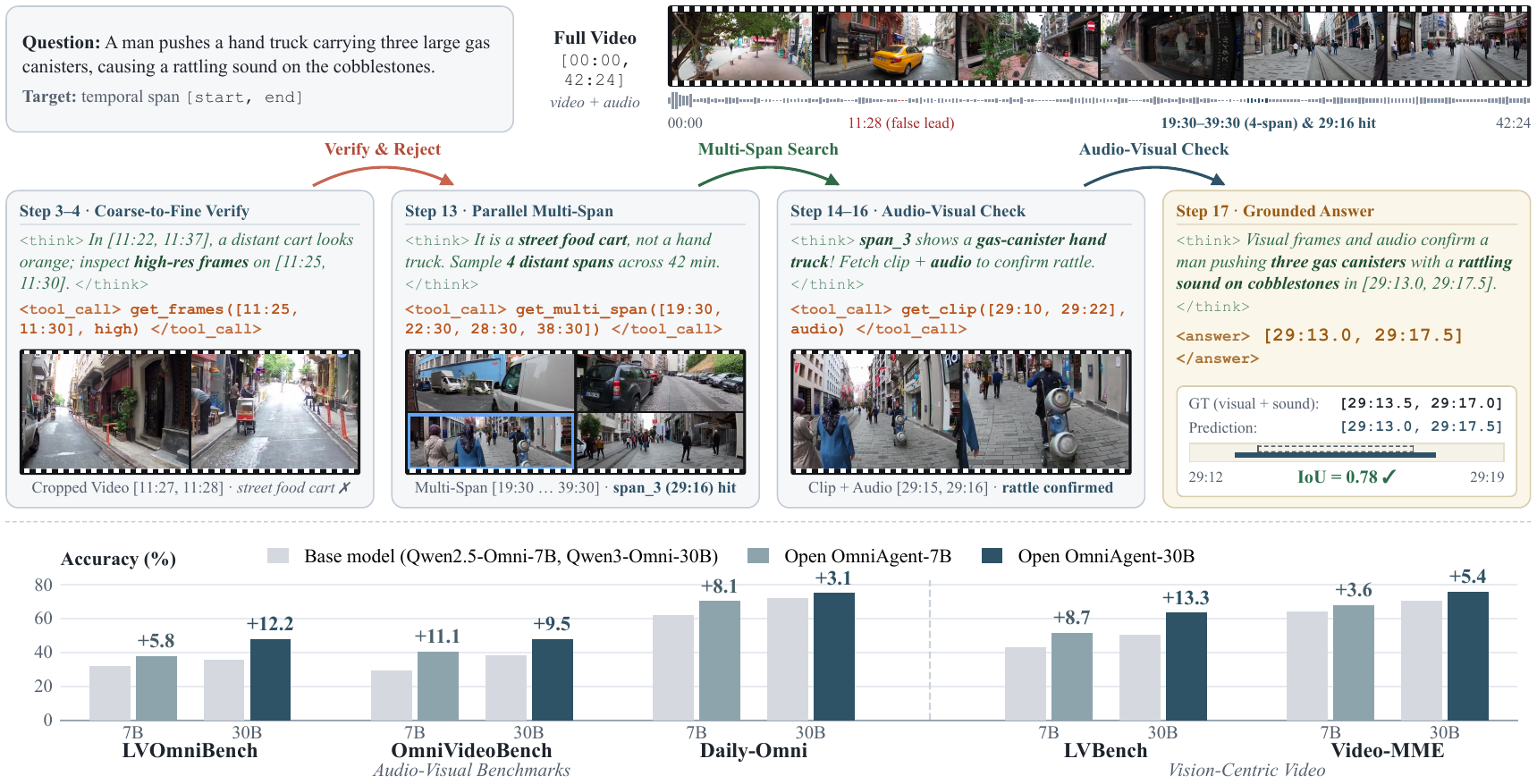}
\caption{\textbf{Open-OmniCoach at a glance.}
\textbf{Top:} A 17-step audio-visual temporal grounding trajectory over a 42-minute video. The agent inspects high-resolution frames to reject a false-lead street food cart (11:28), queries four distant intervals in parallel via \texttt{get\_multi\_span}, verifies both the gas-canister hand truck and cobblestone rattling sound via \texttt{get\_clip}, and grounds the event to \texttt{[29:13.0, 29:17.5]} (\texttt{IoU = 0.78}).
\textbf{Bottom:} Accuracy (\%) of Open OmniAgent-7B and Open OmniAgent-30B compared with their base models (Qwen2.5-Omni-7B and Qwen3-Omni-30B) across three audio-visual and two vision-centric video benchmarks.}
\label{fig:teaser}
\end{figure}

Concretely, we present a three-stage pipeline for building our \textbf{OmniCoach-240K}.
For stage 1, we build an evidence-enhanced Omni-QA pipeline. In particular, we present four steps: fine-grained evidence generation, event-and-evidence matching, skill-guided QA construction, and quality-aware filtering.
Each step runs sequentially, producing refined QA data with fine-grained evidence.
The first stage is a key step in generating a high-quality trajectory.
For stage 2, we design a pipeline for multi-span trajectory generation, using a dynamic omni-evidence harness and omni-evidence data from stage 1 as input. 
We obtain 230.7K multi-span trajectories in this step (Fig.~\ref{fig:teaser}, top).
For stage 3, we present a reflection trajectory generation process. 
We check failed rollouts of the Stage-2 model with AV-decoupled stepwise verification and let the teacher resume from the first confirmed error, yielding an additional 17.4K reflection trajectories (totaling 248.2K trajectories in OmniCoach-240K; Fig.~\ref{fig:data_pipeline}(c)). 
We list the more detailed dataset comparison in Tab.~\ref{tab:dataset_comparison}.
%


%

Building on these data, we train omni-models in two stages.
We first use both multi-span and reflection trajectory data for the SFT process.
For the RL process, as sparse outcome rewards provide limited guidance on intermediate evidence acquisition, we introduce {Step-Level Evidence-Guided Policy Optimization (SEPO)}, which uses the key-evidence annotations to credit each step by its incremental gain in evidence coverage and localization.
Compared with GSPO, our method dynamically assigns credits to each trajectory during each step, leading to stronger performance.

Applying this recipe to open-source models (Qwen3-Omni-30B-A3B~\citep{xu2025qwen3omni} and Qwen2.5-Omni-7B~\citep{xu2025qwen25omni}) yields \textbf{Open OmniAgents, with both 30B and 7B models} (Fig.~\ref{fig:teaser}, bottom).
Empirically, our Open OmniAgent-30B improves over its base model on all five audio-visual benchmarks, raising their average by 5.41 points, including +9.50 on OmniVideoBench~\citep{li2025omnivideobench} and +12.23 on LVOmniBench~\citep{tao2026lvomnibench}.
It also improves other video understanding benchmarks, e.g., +10.28 on Video-MME v2~\cite{fu2026videommev2} and +13.26 on LVBench~\citep{wang2024lvbench}, surpassing Qwen3-VL-235B-A22B on Video-MME v2.
To verify our framework's generalization, we also train Open OmniAgent-7B from Qwen2.5-Omni-7B, which improves the audio-visual benchmarks by an average of 6.25 points.
Moreover, our tasks elicit more varied evidence-seeking behaviors than OmniVideo-100K~\citep{cai2026omnivideo100k} at comparable video length, with 7.1 versus 5.1 effective tool-use patterns.

\section{OmniCoach-240K}
\label{sec:data_synthesis}

\begin{table*}[t]
\centering
\caption{
Comparison of representative video and omni-video training datasets across QA diversity,
dynamic evidence acquisition, and error recovery.
TG: temporal grounding; MCQ: multiple-choice QA; FF: free-form QA/generation;
N.R.: not reported; --: not applicable.
For ours, Avg. Duration is the mean duration of unique videos covered by QA data.
Multi-span denotes multiple temporal intervals requested in one tool call.
In Error Recovery, $\checkmark^\dagger$ denotes synthetic self-reflection samples~\citep{zeng2026videoo3} or teacher retries on environment errors~\citep{xing2026omniagent}, whereas Open OmniCoach recovers from failed student rollouts (Sec.~\ref{subsec_data_stage3}).
}
\label{tab:dataset_comparison}
\resizebox{\textwidth}{!}{
\begin{tabular}{lcccccccc}
\toprule
& &
\multicolumn{4}{c}{\textbf{Diverse QA}} &
\multicolumn{2}{c}{\textbf{Dynamic Evidence Acquisition}} &
\multicolumn{1}{c}{\textbf{Recovery Supervision}} \\
\cmidrule(lr){3-6}
\cmidrule(lr){7-8}
\cmidrule(lr){9-9}

\textbf{Dataset}
& \textbf{Mod.}
& \textbf{\#QA}
& \textbf{Formats}
& \textbf{Avg. Duration}
& \textbf{Evidence Ann.}
& \textbf{\#Traj.}
& \textbf{Multi-span}
& \textbf{Error Recovery} \\

\midrule

LongVALE~\citep{geng2024longvale}
& AV
& N.R.
& TG, FF
& 3.9 min
& Time + text
& --
& --
& $\times$ \\

OmniVideo-100K~\citep{cai2026omnivideo100k}
& AV
& 100K
& MCQ, FF
& 103 s
& Time + text
& --
& --
& $\times$ \\

LongVT~\citep{yang2025longvt}
& V
& N.R.
& TG, FF
& N.R.
& Time
& 34.6K
& $\times$
& $\times$ \\

Video-o3~\citep{zeng2026videoo3}
& V
& N.R.
& MCQ, FF
& N.R.
& Time
& 173K
& $\times$
& $\checkmark^\dagger$ \\

OmniAgent~\citep{xing2026omniagent}
& AV
& N.R.
& TG, MCQ, FF
& N.R.
& N.R.
& 58K
& $\times$
& $\checkmark^\dagger$ \\

\midrule

\textbf{Open OmniCoach (Ours)}
& \textbf{AV}
& \textbf{95.7K}
& \textbf{TG, MCQ, FF}
& \textbf{8.87 min}
& \textbf{Time + modality}
& \textbf{248.2K}
& $\mathbf{\checkmark}$
& $\mathbf{\checkmark}$ \\

\bottomrule
\end{tabular}
}

\vspace{-1em}

\end{table*}

In this section, we present the detailed pipeline for creating our OmniCoach-240K dataset. 
The entire process contains three steps: Hierarchical Evidence QA Generation (Stage-1), Multi-Span Trajectory Generation (Stage-2), and Reflection Trajectory Generation (Stage-3). 
Each stage builds on the output of the previous stage.
The final trajectory data from Stage-2 (230.7K) and Stage-3 (17.4K) together constitute the 248.2K trajectories of OmniCoach-240K.
Appendix~\ref{app:dataset_composition} traces source videos and QA samples through the three stages.


\subsection{Stage-1: Hierarchical Evidence QA Generation}
\label{subsec:task_synthesis}

\begin{figure}[!t]
\centering
\includegraphics[width=\linewidth]{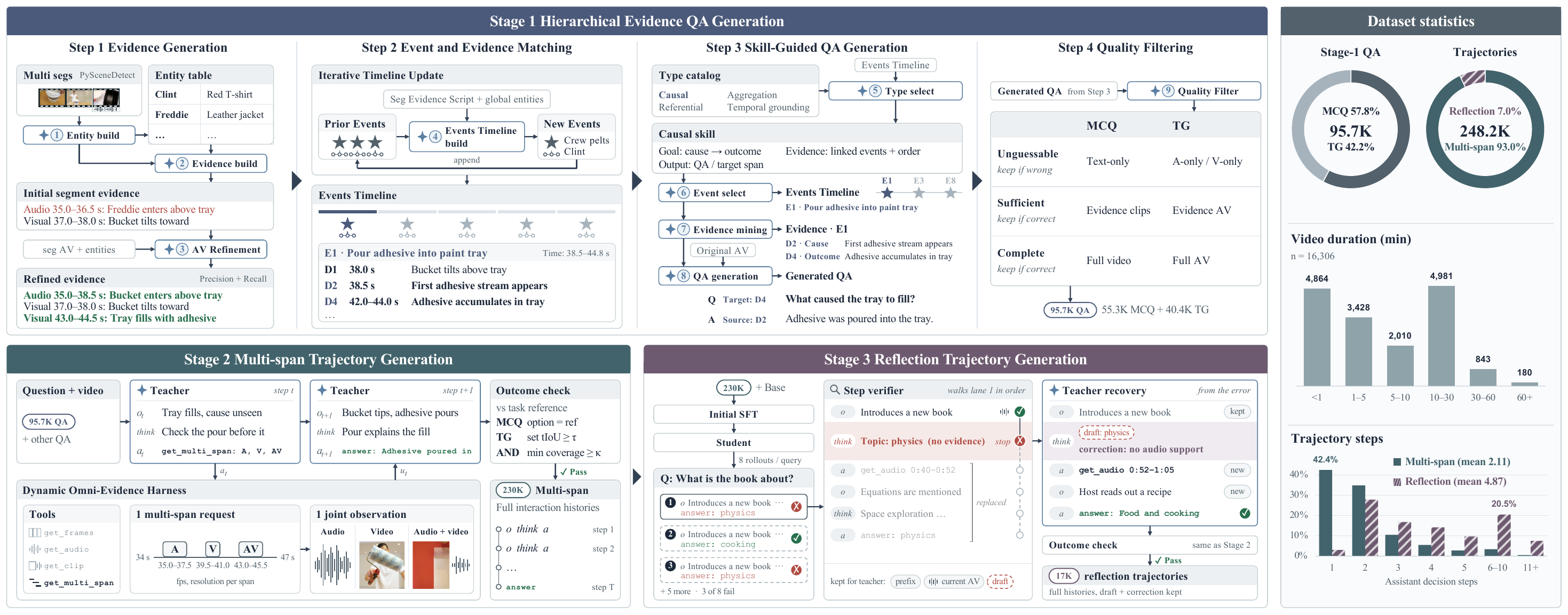}
\caption{
\textbf{OmniCoach-240K data generation pipeline.}
\textbf{(a) Stage 1:} hierarchical audio-visual scripts and global event timelines drive skill-guided MCQ/TG synthesis and verification.
\textbf{(b) Stage 2:} a teacher generates parallel multi-span, multimodal trajectories in the Dynamic Omni-Evidence Harness.
\textbf{(c) Stage 3:} a stepwise verifier locates errors in student rollouts, where the teacher inserts an evidence-grounded correction and continues to an outcome-verified answer.
Right: dataset statistics across QA types, video durations, and trajectory lengths.
}
\label{fig:data_pipeline}
\end{figure}

As shown in Fig.~\ref{fig:data_pipeline} (a), to obtain better Omni QA data, we present a new way of generating hierarchical evidence, which covers fine-grained grounded evidence across video and audio. This stage has four steps: Hierarchical Evidence (step-1), Event and Evidence Matching (step-2), Skill-guided QA Generation (step-3), and Quality-aware Filtering (step-4).


\textbf{Step-1: Fined-grained Evidence Generation.}
We first construct segment-level audio-visual scripts that record fine-grained evidence with time intervals and consistent entity references. Specifically, we divide each video into segments based on visual scene changes and audio pauses, and process context blocks of adjacent segments in temporal order to incrementally build a global entity table of people, sound sources, and objects. Using this table, the model generates a script for each segment, recording fine-grained facts from visual content, speech, music, and other sounds.
To correct errors and recover omitted information in the initial scripts, we introduce \textit{Audiovisual Script Refinement}. Models verify specific facts more reliably than they describe complex segments from scratch~\citep{rawal2025argus,liu2026motionatlas}. The model therefore uses the initial scripts and the entity table as references to re-examine the original audio-visual segments, refine event boundaries, correct sound-source attribution and audio-visual mismatches, fix factual errors, and add omitted information. 

\textbf{Step-2 Event and Evidence Matching.} 
The refined scripts provide rich local details, but selecting task-relevant events and evidence directly from the full record is challenging.
We therefore organize the scripts into a concise global event timeline.
Using the entity table and existing event records, we integrate the scripts in temporal order, merging continuing events and adding new events and state changes.
Each event retains links to its source evidence in the scripts, forming an evidence hierarchy built bottom-up that supports top-down task construction: first selecting events on the timeline, then expanding local details as needed.
Appendix~\ref{app:evidence_construction} details script generation, refinement, and timeline construction.

\textbf{Step3: Skill-guided QA Generation.} %
Building on this evidence hierarchy, we define a skill for each MCQ reasoning type and for TG, specifying the required audio-visual cues, their relations, and how to construct the question.
For example, counting requires verifying candidate events individually and aggregating those that satisfy the question's conditions, while TG targets intervals where the specified audio and visual conditions overlap in time.
A skill catalog summarizes the reasoning goals and evidence requirements of each question type, serving as a concise index of the corresponding skills.

During construction, the model uses the global event timeline and skill catalog to select question types with sufficient supporting evidence and loads the corresponding skills.
It then selects event combinations that satisfy the skill requirements and expands the linked fine-grained scripts to extract both answer-relevant facts and cues that distinguish easily confused entities or events.
Using this evidence together with the corresponding raw audio-visual segments, the model generates QA questions and reference answers, or TG queries and target intervals.
Each task retains the evidence supporting its answer or target intervals as task-level evidence annotations, used for quality filtering and reinforcement learning (Sec.~\ref{subsec:evidence_guided_rl}).
Appendix~\ref{app:task_synthesis_details} provides the complete task skills and construction details for each task type.

\textbf{Step-4: Quality-aware Filtering.} 
For QA, we first remove samples that can be answered correctly from the question and options alone~\citep{chen2024mmstar}.
A verifier model then answers each question in two independent passes: using only the raw audio-visual segments corresponding to the annotated evidence to check whether this evidence suffices to support the reference answer, and using the full video with audio to check whether omitted events outside these segments could change the answer.
We retain only samples for which both answers match the reference answer.
TG follows the same verification principle, requiring a one-to-one match between predicted and target intervals with no missing or extra intervals, and a temporal IoU above a preset threshold for every matched pair.
Appendices~\ref{app:question_answering} and~\ref{app:temporal_grounding} detail the verification settings.

\subsection{Stage-2: Multi-Span Trajectory Generation}
\label{subsec_data_stage2}
As shown in Fig.~\ref{fig:data_pipeline}(b), using the QA data constructed in Stage-1 together with QA samples from other sources, we generate interaction trajectories for training active audio-visual evidence acquisition.
Audio-visual reasoning often requires jointly analyzing evidence distributed across time spans and modalities.
A single query can refer to several disjoint moments in a video, as formalized in one-to-many temporal grounding~\citep{xu2026onetomany}. In our harness, the agent can retrieve several candidate spans in one call and compare their audio-visual evidence before answering.
Retrieving these spans through sequential requests not only increases interaction rounds but may also propagate intermediate misinterpretations or omissions into subsequent decisions, leading to error accumulation.
We therefore support flexible multi-span, multimodal evidence acquisition, allowing the model to directly compare relevant evidence while reducing unnecessary sequential interactions.

To this end, we design the \textit{Dynamic Omni-Evidence Harness}, a multi-turn environment in which the model issues evidence requests and receives corresponding observations until it submits a final answer.
Each request specifies one or more time spans and independently assigns an observation modality to each: audio (A), visual frames (V), or joint audio-visual content (AV).
The harness returns the requested content together as a single observation.
Compared with sequential single-span requests, multi-span requests reduce the rounds needed to retrieve the same set of spans, allowing the model to compare evidence across spans and check audio-visual correspondence within the same turn.
The model can also adjust visual sampling density and spatial resolution as needed.
Additionally, we adopt a cross-turn text memory mechanism similar to OmniAgent~\citep{xing2026omniagent}, retaining key information from previous observations for subsequent reasoning.
Appendix~\ref{app:tool_interface} details the request format and observation configuration.

Given only the task query, initial media overview, and tool interface (without reference answers or evidence annotations), Gemini 3.6 Flash generates its \texttt{observation} and \texttt{think} fields on the fly at each step before invoking a tool or submitting an answer, and we filter the resulting trajectories through an outcome check.
For MCQ, the final selected option must be correct.
For TG, both the set-level temporal intersection over union (tIoU) between predicted and target intervals and the minimum coverage across target intervals must meet their respective thresholds.
We retain 230.7K trajectories for the initial SFT of the student model.

\subsection{Stage-3: Reflection Trajectory Generation}
\label{subsec_data_stage3}

Even after SFT, the student may still produce perceptual hallucinations or make erroneous decisions during interaction.
Learning from successful trajectories alone does not necessarily equip the model to recognize and recover from its own errors.
We therefore construct reflection trajectories from student failures: at confirmed error points, we introduce corrections grounded in audio-visual evidence and continue the interaction to a final answer, providing explicit supervision for error recovery.

\textbf{Trajectory collection.}
We collect student rollouts in the same harness, sampling eight trajectories per query in the QA set and selecting failed trajectories for error diagnosis.

\textbf{Error diagnosis.}
The verifier first selects, from the textual trajectory, key steps that affect the answer or subsequent evidence acquisition and examines them in temporal order.
At each step, it extracts new or revised factual claims relative to the preceding history.
Using the current observations and interaction history, it checks the audio-visual support for these claims and the student's behavioral decisions, such as whether it ignores available counter-evidence.
Verification stops at the first error or inconclusive judgment; only a confirmed error defines a recovery point.

\textbf{Reflection generation.}
Because the diagnosed errors come from failed student rollouts rather than the teacher itself, at each confirmed recovery point we supply the teacher (Gemini 3.6 Flash) with the preceding interaction history, current media observations, the unexecuted failed student turn, and the verifier's diagnostic note.
Preserving the student's erroneous draft, the teacher writes a media-grounded correction and a revised action on the fly, forming an observation $\rightarrow$ erroneous draft $\rightarrow$ correction $\rightarrow$ action step; the diagnostic note is stripped from the exported training data (Appendix~\ref{app:trajectory_refinement}).
The teacher then continues the interaction to a final answer, replacing the original trajectory's remaining steps.
We retain recoveries that pass the same task-level outcome check used in Stage-2, yielding 17.4K reflection trajectories.
The Stage-2 teacher trajectories initialize the student through SFT, while the reflection trajectories provide additional supervision for error correction and recovery in subsequent SFT.

\section{Open Omni Post-training}
\label{sec:agentic_post_training}

\subsection{Supervised Fine-Tuning}
\label{subsec:sft}
We present two-round supervised fine-tuning.
We first train the base model on Stage-2 multi-span trajectories with standard next-token prediction. 
Then, we use the Stage-3 strategy (Sec.~\ref{subsec_data_stage3}) to turn the model’s failed rollouts into reflection trajectories. 
We finally train the base model on both trajectory sets and use the resulting model as the start of reinforcement learning (Sec.~\ref{subsec:evidence_guided_rl}).

\subsection{Reinforcement Learning with Step-Level Credit Assignment}
\label{subsec:evidence_guided_rl}

\textbf{Base Algorithm and Reward Design.} For RL, we follow Open-o3 Video~\citep{meng2025openo3video} and use Group Sequence Policy Optimization (GSPO)~\citep{zheng2025gspo} as the base optimization framework to further optimize the model's multi-turn evidence-acquisition and answering strategies. 
Given an audio-visual input and a question \(x\), the reward for an interaction trajectory \(\tau\) contains two items: outcome reward  \(r_{\mathrm{out}}\)and format reward \(r_{\mathrm{fmt}}\),$R(x,\tau)=r_{\mathrm{out}}(x,\tau)+r_{\mathrm{fmt}}(\tau).
$.
For the multiple-choice question (MCQ) task, the outcome reward is a binary correctness reward based on answer-option matching. 
For the temporal grounding (TG) task, we follow TimeLens2~\citep{zhu2026timelens2} to compute the outcome reward.
The format reward is determined by whether the final answer follows the required output format.
%
Then, GSPO computes a group-relative advantage \(A_i\) for each sampled trajectory. Let
\begin{equation}
\ell(s,A)=
\min\left\{
sA,\,
\operatorname{clip}(s,1-\epsilon,1+\epsilon)A
\right\}
\end{equation}
denotes the clipped objective term. The GSPO objective is
\begin{equation}
\mathcal J_{\mathrm{GSPO}}(\theta)
=
\mathbb E_{x,\{\tau_i\}}
\left[
\frac{1}{G}
\sum_{i=1}^{G}
\ell\!\left(s_i(\theta),A_i\right)
\right].
\end{equation}
Here, \(\{\tau_i\}_{i=1}^{G}\) are \(G\) trajectories sampled by the old policy for the same input, \(s_i(\theta)\) is the length-normalized sequence importance ratio, and \(\epsilon\) is the clipping threshold.

\begin{wrapfigure}{r}{0.46\linewidth}
\vspace{-0.4em}
\centering
\includegraphics[width=\linewidth]{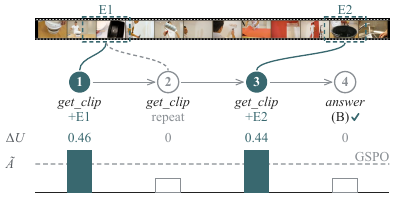}
\vspace{-0.4em}
\caption{SEPO on one rollout (illustrative values). A step gets $\Delta U>0$ only when its retrieved segment covers new annotated evidence ($E_1$, $E_2$). GSPO gives every step the same advantage (dashed line); SEPO scales it by the gain without changing its sign.}
\label{fig:sepo}
\vspace{-0.4em}
\end{wrapfigure}
\textbf{Step-Level Evidence Guidance.}
A shared trajectory-level advantage cannot distinguish the evidence contribution of individual steps: successful trajectories may contain redundant calls, while failed trajectories may still acquire useful evidence. 
Ours MCQ and TG data provide evidence temporal annotations, which offer a common reference for evaluating evidence acquisition. 
We therefore propose {Step-Level Evidence-Guided Policy Optimization (SEPO)}, which uses evidence-matching gains to adjust the update strength of each step (Fig.~\ref{fig:sepo}).

Let \(\mathcal A_x\) denote the annotated evidence intervals, and let \(\mathcal C_{i,k}\) denote all candidate segments actually retrieved by trajectory \(\tau_i\) up to step \(k\). 
To prevent irrelevant candidates from affecting the score of the evidence already acquired, the scorer performs a bounded exhaustive subset search over \(\mathcal C_{i,k}\) and denotes the highest-scoring subset by \(\mathcal S^\star_{i,k}\):
\begin{equation}
\mathcal S^\star_{i,k}
\in
\arg\max_{\mathcal S\subseteq\mathcal C_{i,k}}
q(\mathcal S,\mathcal A_x).
\end{equation}
This search is used only to compute the evidence utility and does not represent an additional model action.
The current evidence utility is
\begin{equation}
\widehat U_{i,k}
=
q(\mathcal S^\star_{i,k},\mathcal A_x).
\end{equation}
We maintain the historical best utility and use its improvement as the evidence-matching gain of the current step:
\begin{equation}
U_{i,k}
=
\max\left(U_{i,k-1},\widehat U_{i,k}\right),
\qquad
\Delta U_{i,k}
=
U_{i,k}-U_{i,k-1},
\qquad
U_{i,0}=0.
\end{equation}

We standardize and clip the gains within each trajectory, and define the step-level advantage as
\begin{equation}
\widetilde A_{i,k}
=
A_i
+
\lambda |A_i|
\operatorname{clip}\!\left(
\frac{\Delta U_{i,k}-\mu_i}{\sigma_i+\delta},
-1,1
\right).
\end{equation}
Here, \(\mu_i\) and \(\sigma_i\) are the mean and standard deviation of the step gains within trajectory \(i\), \(\delta>0\) is a numerical stability term, and \(0\leq\lambda<1\) controls the adjustment strength. 
This adjustment preserves the sign of the advantage, giving high-gain steps stronger reinforcement in positive-advantage trajectories and weaker suppression in negative-advantage trajectories.

We adopt the GSPO-token formulation, assign each step advantage to its corresponding generated tokens, and maximize the following objective~\citep{zheng2025gspo}:
\begin{equation}
\mathcal J_{\mathrm{SEPO}}(\theta)
=
\mathbb E_{x,\{\tau_i\}}
\left[
\frac{1}{G}
\sum_{i=1}^{G}
\frac{1}{N_i}
\sum_{t=1}^{N_i}
\ell\!\left(
s_{i,t}(\theta),
\widetilde A_{i,k(i,t)}
\right)
\right].
\end{equation}
Here, \(N_i\) is the number of generated tokens in trajectory \(i\), and \(k(i,t)\) denotes the step to which token \(t\) belongs. 
The value of \(s_{i,t}(\theta)\) is equal to \(s_i(\theta)\), while gradients propagate only through the conditional generation probability of the current token, so \(\widetilde A_{i,k(i,t)}\) scales the gradient contribution of step \(k\)'s tokens. 
In implementation, we fold turn-level protocol penalties for invalid or duplicate calls into the standardized step score before computing \(\widetilde A_{i,k}\) and perform a single on-policy gradient update per rollout batch, under which our turn-segmented trainer yields the exact same token-level policy gradient as the GSPO-token objective above (Appendix~\ref{app:step_reward_optimization}).
Setting \(\lambda=0\) recovers our GSPO baseline under the identical codebase, trajectory rewards, protocol checks, and KL regularization.

\makeatletter
\newcommand{\papertablestyle}{%
  \fontencoding{T1}\selectfont
  \fontsize{8.5}{10.5}\selectfont
  \setlength{\tabcolsep}{3pt}%
  \renewcommand{\arraystretch}{1.18}%
  \sisetup{detect-weight=true,detect-family=true,table-number-alignment=center,retain-explicit-plus=true}%
  \setlength{\abovecaptionskip}{0pt}%
  \setlength{\belowcaptionskip}{6pt}}
\makeatother

\section{Experiments}
\label{sec:exp}

\subsection{Experimental Setup}
\label{sec:exp_setup}

\paragraph{Evaluation Benchmarks.} We evaluate our method on benchmarks covering omni-video (audio-visual) understanding and general video understanding. WorldSense~\citep{hong2025worldsense}, Daily-Omni~\citep{zhou2025dailyomni}, JointAVBench~\citep{chao2025jointavbench}, OmniVideoBench~\citep{li2025omnivideobench}, and LVOmniBench~\citep{tao2026lvomnibench} assess audio-visual understanding, focusing on the integration of auditory and visual evidence. 
Video-MME~\citep{fu2024videomme}, Video-MME v2~\citep{fu2026videommev2}, LongVideoBench~\citep{wu2024longvideobench}, LVBench~\citep{wang2024lvbench}, and VideoZeroBench~\citep{meng2026videozerobench} assess general video understanding, including temporal reasoning.


\paragraph{Baselines and Comparison Models.} 
We compare against three groups of baselines.
Proprietary models include GPT-4o~\citep{openai2024gpt4o}, Gemini-2.0-Flash\footnote{\url{https://deepmind.google/models/model-cards/}}, Gemini-2.5-Pro~\citep{geminiteam2025gemini25}, and Gemini-3.0-Pro~\citep{tao2026lvomnibench}.
Open-source models contains VideoLLaMA2~\citep{cheng2024videollama2}, Qwen2.5-VL~\citep{bai2025qwen25vl}, Qwen2.5-Omni~\citep{xu2025qwen25omni}, and Qwen3-Omni-30B-A3B-Instruct~\citep{xu2025qwen3omni}.
For agentic models, we compare with OmniAgent~\citep{xing2026omniagent} and Video-o3~\citep{zeng2026videoo3}.

\paragraph{Training details and inference.}
We build our agent on Qwen3-Omni-30B-A3B-Instruct~\citep{xu2025qwen3omni}. 
%
%
As discussed in Sec.~\ref{sec:agentic_post_training}, we first train on 230.7K Stage-2 multi-span trajectories. 
Then, we combine these 230.7K trajectories with 17.4K Stage-3 reflection trajectories, yielding 248.2K training trajectories. 
Both supervised training runs use a learning rate of $10^{-5}$ for two epochs. 
We use AdamW as the optimizer. For RL training, we adopt our SEPO, with 8 rollouts per task and a learning rate of $10^{-6}$ for 834 updates. 
At inference, we allow up to 12 interaction turns. 
We also train Open OmniAgent-7B from Qwen2.5-Omni 7B~\citep{xu2025qwen25omni} with the same recipe.

\subsection{Main Results}
\label{sec:exp_main}



\begin{table}[!t]
\centering
\papertablestyle
\caption{Results on Omni-Video understanding benchmarks. All scores are accuracy (\%). Bold and underline mark the best and second-best open-source results; -- denotes unavailable results. Avg.\ is the mean over the five benchmarks, and $\Delta$ is computed from unrounded scores against the reported scores of each base model (Qwen2.5-Omni for Open OmniAgent-7B and Qwen3-Omni for Open OmniAgent-30B) before rounding.}
\label{tab:main_results}
\setlength{\tabcolsep}{0pt}
\sisetup{mode=text,table-format=+2.1}
\begin{tabularx}{\linewidth}{@{}>{\raggedright\arraybackslash}X >{\centering\arraybackslash}p{40pt} S[table-column-width=48pt] S[table-column-width=48pt] S[table-column-width=58pt] S[table-column-width=66pt] S[table-column-width=56pt] S[table-column-width=32pt]@{}}
\toprule
\textbf{Method} & \textbf{Size} & {\hd{WorldSense}} & {\hd{Daily-Omni}} & {\hd{JointAVBench}} & {\hd{OmniVideoBench}} & {\hd{LVOmniBench}} & {\hd{\textbf{Avg.}}} \\
\midrule
\grp{8}{Proprietary Models} \\
GPT-4o & \sz{--} & 42.6 & 56.5 & 45.0 & {--} & {--} & {--} \\
Gemini-2.0-Flash & \sz{--} & {--} & 67.8 & {--} & 41.5 & 42.9 & {--} \\
Gemini-2.5-Pro & \sz{--} & 65.1 & 80.6 & 65.3 & 58.9 & {--} & {--} \\
Gemini-3.0-Pro & \sz{--} & 66.4 & 81.1 & {--} & 61.8 & 65.8 & {--} \\
\midrule
\grp{8}{Open-source Generalist Models} \\
video-SALMONN 2+ & \sz{7B} & 50.9 & 71.8 & 46.7 & 36.4 & 32.7 & 47.7 \\
OmniVinci & \sz{9B} & 48.2 & 66.5 & {--} & 32.1 & {--} & {--} \\
Qwen2.5-VL & \sz{7B} & 35.5 & 40.7 & 47.7 & 29.8 & {--} & {--} \\
Qwen2.5-Omni & \sz{7B} & 45.4 & 62.1 & 56.5 & 29.3 & 32.0 & 45.1 \\
Qwen3-Omni-Instruct & \sz{30B-A3B} & {\underline{54.0}} & {\underline{71.9}} & {\underline{63.6}} & 38.4 & 35.8 & {\underline{52.7}} \\
\midrule
\grp{8}{Open-source Agentic Models} \\
OmniAgent & \sz{7B} & 47.2 & 64.8 & {--} & 37.1 & {--} & {--} \\
\midrule
\rowcolor{OursRow}
\textbf{Open OmniAgent (Ours)} & \sz{7B} & 48.9 & 70.2 & 59.3 & {\underline{40.4}} & {\underline{37.8}} & 51.3 \\
\quad\textit{$\Delta$ vs.\ Qwen2.5-Omni} & & \dpos +3.5 & \dpos +8.1 & \dpos +2.8 & \dpos +11.1 & \dpos +5.8 & \dpos +6.3 \\
\rowcolor{OursRow}
\textbf{Open OmniAgent (Ours)} & \sz{30B-A3B} & \bfseries 54.8 & \bfseries 75.0 & \bfseries 64.9 & \bfseries 47.9 & \bfseries 48.0 & \bfseries 58.1 \\
\quad\textit{$\Delta$ vs.\ Qwen3-Omni} & & \dpos +0.8 & \dpos +3.2 & \dpos +1.3 & \dpos +9.5 & \dpos +12.2 & \dpos +5.4 \\
\bottomrule
\end{tabularx}
\end{table}

\begin{table}[!t]
\centering
\papertablestyle
\caption{Results on vision-centric video understanding benchmarks. Scores are accuracy (\%) except for the official Video-MME v2 Rating; for VideoZeroBench (VZB), we report Level-3 accuracy. Video-MME v2 baselines use the with-subtitle/audio setting of its leaderboard. Other conventions follow Tab.~\ref{tab:main_results}.}
\label{tab:main_video}
\setlength{\tabcolsep}{0pt}
\sisetup{mode=text,table-format=+2.1}
\begin{tabularx}{\linewidth}{@{}>{\raggedright\arraybackslash}X >{\centering\arraybackslash}p{42pt} S[table-column-width=48pt] S[table-column-width=36pt] S[table-column-width=38pt] S[table-column-width=60pt] S[table-column-width=46pt] S[table-column-width=36pt]@{}}
\toprule
\lowhd{\textbf{Method}} & \lowhd{\textbf{Size}} & {\lowhd{\hd{Video-MME}}} & \multicolumn{2}{c}{\hd{Video-MME v2}} & {\lowhd{\hd{LongVideoBench}}} & {\lowhd{\hd{LVBench}}} & {\hd{VZB}} \\
\cmidrule(lr){4-5}
 & & & {\hd{Acc}} & {\hd{Rating}} & & & {\hd{L3}} \\
\midrule
\grp{8}{Proprietary Models} \\
GPT-4o & \sz{--} & 71.9 & {--} & {--} & 66.7 & 48.9 & {--} \\
Gemini-1.5-Pro & \sz{--} & 75.0 & {--} & {--} & 64.0 & 33.1 & {--} \\
Gemini-2.5-Pro & \sz{--} & {--} & {--} & {--} & 76.8 & 67.4 & 16.4 \\
Gemini-3.0-Pro & \sz{--} & 88.6 & 66.1 & 49.4 & 75.9 & 77.0 & 17.0 \\
\midrule
\grp{8}{Open-source Generalist Models} \\
Qwen2.5-Omni & \sz{7B} & 64.3 & {--} & {--} & 53.8 & 43.0 & {--} \\
Qwen3-Omni-Instruct & \sz{30B-A3B} & 70.5 & 34.1 & 17.1 & {\underline{66.9}} & 50.2 & {--} \\
Qwen3-VL-Instruct & \sz{8B} & 71.4 & 35.7 & 18.2 & 66.4 & 58.0 & 8.2 \\
Qwen3-VL-Instruct & \sz{235B-A22B} & {\underline{79.2}} & {\underline{43.3}} & {\underline{25.0}} & {--} & {\underline{67.7}} & 9.6 \\
Qwen3.5 & \sz{397B-A17B} & \bfseries 83.7 & 42.1 & 24.5 & {--} & \bfseries 75.5 & {\underline{10.6}} \\
\midrule
\grp{8}{Open-source Agentic Models} \\
OmniAgent & \sz{7B} & 67.8 & {--} & {--} & {--} & 50.5 & {--} \\
\midrule
\rowcolor{OursRow}
\textbf{Open OmniAgent-7B} & \sz{7B} & 67.9 & 36.5 & 18.4 & 57.1 & 51.7 & 6.4 \\
\quad\textit{$\Delta$ vs.\ Qwen2.5-Omni} &  & \dpos +3.6 & & & \dpos +3.3 & \dpos +8.7 & \\
\rowcolor{OursRow}
\textbf{Open OmniAgent-30B} & \sz{30B-A3B} & 75.9 & \bfseries 44.4 & \bfseries 25.5 & \bfseries 67.4 & 63.5 & \bfseries 12.0 \\
\quad\textit{$\Delta$ vs.\ Qwen3-Omni} &  & \dpos +5.4 & \dpos +10.3 & \dpos +8.4 & \dpos +0.5 & \dpos +13.3 & \\
\bottomrule
\end{tabularx}
\end{table}

\noindent
\textbf{Results on omni-video understanding benchmarks.}
As shown in Tab.~\ref{tab:main_results}, our Open OmniAgent-30B achieves the best reported performance among the compared open-source models on all five audio-visual benchmarks, with an average accuracy of 58.1\%, exceeding the reported Qwen3-Omni result by 5.4 percentage points.
The largest gains are observed on OmniVideoBench and LVOmniBench, reaching 9.5 and 12.2 percentage points, respectively.
These gains are consistent with our interaction-oriented supervision, which trains agents to acquire and integrate task-relevant evidence through multi-turn interactions.
The improvements also extend to Qwen2.5-Omni-7B: Open OmniAgent-7B achieves an average accuracy of 51.3\%, improving over the reported base-model result by 6.3 percentage points. 
These consistent gains across model backbones support the transferability of our training data and recipe.

\noindent
\textbf{Results on general video understanding benchmarks.}
Open OmniAgent-30B also achieves competitive results on general video understanding benchmarks (Tab.~\ref{tab:main_video}), improving over the reported Qwen3-Omni results by 10.3 and 13.3 percentage points in accuracy on Video-MME v2 and LVBench, respectively.
These results suggest that learning to acquire and integrate task-relevant evidence through interaction also benefits broader video understanding, beyond dedicated audio-visual benchmarks.

\subsection{Ablation Studies and Analysis}
\label{sec:exp_data}

\textbf{Scaling Effect of Open-OmniCoach.}
The OmniVideo-100K~\citep{cai2026omnivideo100k} provides question--answer pairs, whereas our data additionally provides interaction trajectories.
We therefore train on each dataset in the supervision format it provides: direct SFT, which trains the model to answer from the input media without tool calls, for OmniVideo-100K, and agentic SFT, which trains the model on multi-turn interaction trajectories, for our data.
We vary the training data from 25\% to 100\% of each dataset and report the mean and per-benchmark accuracy on WorldSense, Daily-Omni, OmniVideoBench, and LVOmniBench in Appendix Fig.~\ref{fig:data_scaling} and Tab.~\ref{tab:data_scaling_full}.
Overall, our results show that our data pipeline scales better and performs more effectively than OmniVideo-100k.

\textbf{Effectiveness of Query Set in Open-OmniCoach.}
Our query set includes long-video tasks and offers broader task diversity (Tab.~\ref{tab:dataset_comparison}).
To evaluate its value for agent training, we compare training on 5K trajectories generated from each query source using the same procedure.
Using our queries instead of OmniVideo-100K queries improves accuracy across all four benchmarks by an average of 1.0 percentage point (Appendix Tab.~\ref{tab:query_source}).

\begin{table}[!t]
\centering
\papertablestyle
\renewcommand{\arraystretch}{1.12}
\caption{Ablation studies on Qwen3-Omni-30B-A3B-Instruct. SFT variants train from the base model; RL runs start from + Reflection Trajectory. Accuracy (\%) $\uparrow$.}
\label{tab:ablation}
\setlength{\tabcolsep}{0pt}
\sisetup{mode=text,table-format=2.1}
\begin{tabularx}{\linewidth}{@{}>{\raggedright\arraybackslash}X >{\centering\arraybackslash}p{48pt} S[table-column-width=50pt] S[table-column-width=60pt] S[table-column-width=66pt] S[table-column-width=56pt] S[table-column-width=32pt]@{}}
\toprule
\lowhd{\textbf{Method / Setting}} & \lowhd{\hd{\textbf{Stage}}} & \multicolumn{2}{c}{\hd{\textbf{Video understanding}}} & \multicolumn{2}{c}{\hd{\textbf{Audio-visual reasoning}}} & {\lowhd{\hd{\textbf{Avg.}}}} \\
\cmidrule(lr){3-4}\cmidrule(lr){5-6}
 & & {\hd{Video-MME}} & {\hd{Video-MME-v2}} & {\hd{OmniVideoBench}} & {\hd{LVOmniBench}} & \\
\midrule
Qwen3-Omni-30B-A3B-Instruct & \sz{Base} & 70.5 & 34.1 & 38.4 & 35.8 & 44.7 \\
+ Multi-Span Trajectory & \sz{SFT} & 72.9 & 40.7 & 44.7 & 46.8 & 51.3 \\
+ Reflection Trajectory & \sz{SFT} & 74.4 & 42.6 & 46.3 & 47.5 & 52.7 \\
\midrule
+ GSPO (RL baseline) & \sz{RL} & 75.4 & 43.3 & 46.9 & 47.7 & 53.3 \\
\rowcolor{OursRow}
\textbf{+ SEPO (Ours, final)} & \sz{RL} & \bfseries 75.9 & \bfseries 44.4 & \bfseries 47.9 & \bfseries 48.0 & \bfseries 54.1 \\
\bottomrule
\end{tabularx}
\vspace{-0.5em}
\end{table}


%
%
\textbf{Effectiveness of Open-OmniCoach Framework.}
We list the detailed effectiveness of Open-OmniCoach in Tab.~\ref{tab:ablation}.
%
%
%
Starting from the baseline model, adding multi-span trajectories leads to a significant improvement, as shown in the second row.
%
%
With reflection trajectories, we observe more improvements across all four benchmarks in the third row.
For the strong GSPO baseline, adopting RL post-training gives a further 0.6-point average gain across the four benchmarks.
%
%
%
Then, using our proposed SEPO, the further gains add 1.10 and 0.99 points over GSPO on Video-MME-v2 and OmniVideoBench.
Appendix~\ref{app:additional_experiments} further evaluates intermediate RL checkpoints (Tab.~\ref{tab:rl_checkpoints}) and audits process-level evidence acquisition and error recovery on OmniVideoBench (Tab.~\ref{tab:ovb_process_audit}).

\textbf{Controlled Analysis of Reflection Supervision.}
To separate error-recovery supervision from extra trajectory volume, we run a controlled ablation on a 25\% OmniCoach-240K subset (62.0K trajectories: 57.7K multi-span + 4.4K reflection; Tab.~\ref{tab:reflection_ablation}).
Replacing the 4.4K reflection trajectories with 4.4K Stage-2 successful trajectories lowers the four-benchmark mean from 49.2\% to 48.0\% ($-1.5$ on Video-MME-v2 and $-1.3$ on OmniVideoBench).
With pre-mistake student turns masked ($\text{loss}=0$), supervising only the first corrective turn uses fewer tokens per epoch than the equal-volume success baseline (20.94M vs.\ 21.15M) yet reaches 48.5\%; two or three turns reach 48.9\% and 49.1\%, recovering most of the full gain (49.2\%, 22.82M).

\begin{table}[!t]
\centering
\papertablestyle
\renewcommand{\arraystretch}{1.12}
\caption{Controlled ablation of reflection supervision on a 25\% OmniCoach-240K subset (62.0K trajectories total; counts rounded independently) with Qwen3-Omni-30B-A3B-Instruct. Pre-mistake student turns are masked in all reflection variants. Accuracy (\%) $\uparrow$.}
\label{tab:reflection_ablation}
\setlength{\tabcolsep}{0pt}
\sisetup{mode=text,table-format=2.1}
\begin{tabularx}{\linewidth}{@{}>{\raggedright\arraybackslash}X >{\centering\arraybackslash}p{58pt} S[table-column-width=50pt] S[table-column-width=60pt] S[table-column-width=66pt] S[table-column-width=56pt] S[table-column-width=32pt]@{}}
\toprule
\lowhd{\textbf{57.7K Multi-Span +}} & \lowhd{\hd{\textbf{Tokens / Ep.}}} & \multicolumn{2}{c}{\hd{\textbf{Video understanding}}} & \multicolumn{2}{c}{\hd{\textbf{Audio-visual reasoning}}} & {\lowhd{\hd{\textbf{Avg.}}}} \\
\cmidrule(lr){3-4}\cmidrule(lr){5-6}
 & & {\hd{Video-MME}} & {\hd{Video-MME-v2}} & {\hd{OmniVideoBench}} & {\hd{LVOmniBench}} & \\
\midrule
4.4K Stage-2 Success & \sz{21.15M} & 72.0 & 37.4 & 41.4 & 41.3 & 48.0 \\
\midrule
4.4K Reflection (1 turn) & \sz{20.94M} & 72.2 & 37.9 & 42.1 & 41.7 & 48.5 \\
4.4K Reflection (2 turns) & \sz{21.56M} & 72.6 & 38.5 & 42.5 & 42.1 & 48.9 \\
4.4K Reflection (3 turns) & \sz{21.96M} & \bfseries 72.9 & 38.7 & 42.4 & \bfseries 42.5 & 49.1 \\
\rowcolor{OursRow}
\textbf{4.4K Reflection (Full)} & \sz{22.82M} & 72.8 & \bfseries 38.9 & \bfseries 42.7 & 42.4 & \bfseries 49.2 \\
\bottomrule
\end{tabularx}
\vspace{-0.5em}
\end{table}

\section{Related Work}
\label{sec:related_work}

\subsection{From Vision-Language to Omni-Modal Datasets}
DenseWorld-1M~\citep{li2025denseworld1m} builds dense image captions by combining mask-grounded object descriptions with spatial relations between objects. Large-scale video-language datasets have laid the foundation for multimodal video understanding. WebVid~\citep{Bain2021FrozenITwebvid2m}, InternVid~\citep{wang2024internvid}, and Panda-70M~\citep{Chen2024Panda70MC7} scale video-text pretraining with web-collected or automatically generated captions, while Video-ChatGPT~\citep{Maaz2023VideoChatGPTTD} and LLaVA-Video~\citep{Zhang2024LLaVAVideoVI} construct instruction-following supervision for video understanding. Because these video-language datasets remain predominantly vision-centric, they provide limited supervision for joint audio-visual reasoning.

Recent omni-modal datasets incorporate audio alongside visual content to capture cross-modal and temporal interactions. Omni-Captioner~\citep{ma2025omnicaptioner}, TimeChat-Captioner~\citep{yao2026timechatcaptioner}, and AVSCap~\citep{wang2026avscap} organize audio-visual streams into structured descriptions, while OmniVideo-100K~\citep{cai2026omnivideo100k} links entity-anchored scripts into cross-segment evidence chains to synthesize reasoning question-answer pairs. However, these captioning and question-answering datasets supervise only static descriptions or final task answers: even when OmniVideo-100K annotates cross-segment evidence, those annotations are used only to construct question-answer pairs, leaving the multi-turn process of acquiring and verifying evidence unsupervised. In contrast, \textbf{OmniCoach-240K} combines hierarchical task-level evidence annotations with outcome-verified multi-span trajectories and diagnosis-guided reflection trajectories, supervising how agents find, connect, and correct task-relevant audio-visual evidence.

\subsection{From Video MLLMs to Omni Agents}
Vision-language models such as Qwen2.5-VL~\citep{bai2025qwen25vl} establish strong visual perception and reasoning over images and videos, and recent multimodal models extend perception to unified audio-visual inputs. VideoLLaMA2~\citep{cheng2024videollama2} integrates spatial-temporal visual modeling with audio understanding, while Qwen2.5-Omni~\citep{xu2025qwen25omni} and Qwen3-Omni~\citep{xu2025qwen3omni} jointly model visual, acoustic, and linguistic streams for end-to-end audio-visual reasoning.

GAR~\citep{wang2026gar} uses RoI-aligned feature replay to reason about regions in their scene context and model interactions across regions. SAMTok~\citep{zhou2026samtok} represents a segmentation mask with two discrete tokens, enabling autoregressive prediction of pixel-level outputs. For video reasoning, Open-o3-Video~\citep{meng2025openo3video} trains a model to identify timestamps and object bounding boxes as explicit evidence for its answer.

Beyond single-pass perception, recent works formulate video understanding as an active evidence-seeking process. LongVT~\citep{yang2025longvt}, VideoSeek~\citep{lin2026videoseek}, and MedScope~\citep{li2026medscope} retrieve or inspect task-relevant video segments during reasoning. Extending active perception to audio-visual understanding, OmniAgent frameworks~\citep{tao2025omniagent,xing2026omniagent} introduce active audio-visual perception, while AOP-Agent~\citep{xu2026aopagent} targets multi-hop audio-visual reasoning through agentic tool use. Yet existing video and omni-modal agent frameworks focus primarily on agent architectures and interaction mechanisms: their training pipelines offer limited support for flexible multi-span, multimodal evidence acquisition and supervise mainly successful rollouts (or synthetic reflections and environment-error retries), rather than recovering from the student agent's own intermediate perceptual and decision errors. Within \textbf{Open-OmniCoach}, we pair the \textit{Dynamic Omni-Evidence Harness} for multi-span evidence acquisition with multi-span and reflection trajectories derived from diagnosed student failures to train \textbf{Open OmniAgent}.

\subsection{Reinforcement Learning for Multimodal Agents}
Reinforcement learning has increasingly been used to elicit active perception and tool-use behaviors in multimodal models. DeepEyes~\citep{zheng2025deepeyes} learns visual tool invocation through end-to-end reinforcement learning, MMSearch-R1~\citep{wu2025mmsearchr1} optimizes multi-turn multimodal search with outcome-based rewards, and work on traceable visual grounded reasoning~\citep{wang2026traceable} incorporates explicit evidence-grounding signals into reinforcement learning to improve the faithfulness of intermediate reasoning. At the optimization level, GSPO~\citep{zheng2025gspo} provides stable sequence-level policy optimization for large models. When applied to multi-turn evidence seeking, however, sequence-level outcome rewards and static evidence-grounding rewards assign uniform credit across a trajectory rather than distinguishing individual evidence-acquisition steps. Our Step-Level Evidence-Guided Policy Optimization (\textbf{SEPO}) instead computes incremental step-level evidence-matching gains from task-level evidence annotations to adjust trajectory-level advantages across steps, providing finer-grained credit assignment for multi-turn evidence acquisition.

\section{Conclusion}
\label{sec:conclusion}
In this work, we introduced Open-OmniCoach, a framework for training evidence-seeking omni video agents through evidence-enhanced trajectory learning. 
We construct fine-grained audio-visual evidence annotations and use them to generate multi-span interaction trajectories that teach agents to actively acquire and connect evidence across time and modalities. 
We further turn the agent's own failed rollouts into reflection trajectories, providing explicit supervision for recovering from intermediate errors.
Building on these data, we propose Step-Level Evidence-Guided Policy Optimization (SEPO), which uses incremental evidence-matching gains to provide finer-grained credit for individual interaction steps. 
Across both 7B and 30B backbones, Open OmniAgent consistently improves audio-visual reasoning and general video understanding, while our ablations demonstrate the complementary benefits of multi-span trajectories, reflection supervision, and step-level evidence guidance. 
These results highlight the importance of explicitly supervising how omni video agents find, connect, and correct task-relevant evidence.

\section*{Acknowledgments}
We thank Minghua Ma for insightful discussions on the evaluation harness design and data attribution analysis, and Jiahao Meng for helpful discussions on grounded video reasoning and valuable feedback throughout this work.

\section{Contributions}
\label{app:contributions}

\subsection*{Authors}
Weisong Liu$^{1,2}$,
Xiangtai Li$^{1,\ddagger}$,
Haochen Wang$^{2}$,
Haiquan Wen$^{1}$,
Zhaoxiang Zhang$^{2,\dagger}$

\subsection*{Affiliations}
\noindent
$^{1}$ByteDance \\
$^{2}$Institute of Automation, Chinese Academy of Sciences (CASIA)

\vspace{0.5em}
\noindent
$^{\ddagger}$Project leader. \\
$^{\dagger}$Corresponding author.

\vspace{1em}
\bibliographystyle{plainnat}
\bibliography{references}

\clearpage
\beginappendix
\suppressfloats[t]
\noindent
\textbf{Overview.}
Appendix~\ref{app:additional_experiments} reports additional experiments, including temporal grounding, data scaling, query-source comparison, process-level audits on OmniVideoBench, and RL checkpoint progression.
Appendices~\ref{app:task_synthesis_details}--\ref{app:evidence_reward} provide implementation details for task construction, trajectory synthesis and verification, and step-level evidence reward optimization, and Appendix~\ref{app:dataset_composition} summarizes the QA data flow of OmniCoach-240K.
We also discuss the future work of our data pipeline and methods.



\section{Additional Experiments and Analysis}
\label{app:additional_experiments}

\textbf{Temporal grounding.} Table~\ref{tab:main_grounding} reports mIoU on TimeLens and LongVALE. Open OmniAgent-7B achieves 45.9\% on TimeLens and 43.2\% on LongVALE, while Open OmniAgent-30B achieves 49.8\% and 47.5\%, respectively.

\begin{figure}[!t]
\centering
\includegraphics[width=0.52\linewidth]{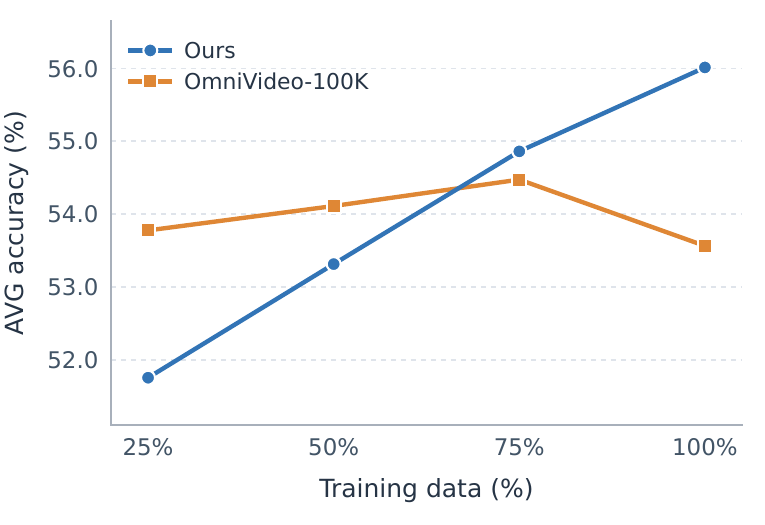}
\caption{Data scaling using 25\%--100\% of OmniVideo-100K and our dataset. Per-benchmark results are in Tab.~\ref{tab:data_scaling_full}.}
\label{fig:data_scaling}
\end{figure}

\textbf{Data scaling and query-source comparison.} Figure~\ref{fig:data_scaling} and Table~\ref{tab:data_scaling_full} report the aggregate and per-benchmark scaling results from 25\% to 100\% training data, and Table~\ref{tab:query_source} compares 5K agentic trajectories synthesized from OmniVideo-100K queries versus our Stage-1 queries under the same pipeline.

\begin{table}[!t]
\centering
\papertablestyle
\caption{Temporal grounding. TimeLens and LongVALE report mIoU (\%). Baseline TimeLens scores are sample-weighted over the three TimeLens-Bench subsets reported by TimeLens2 (Qwen2.5-VL-7B by TimeLens)~\citep{zhu2026timelens2}; LongVALE baselines follow the LongVALE report and OmniAgent~\citep{geng2024longvale,xing2026omniagent}; -- denotes unreported results.}
\label{tab:main_grounding}
\setlength{\tabcolsep}{0pt}
\sisetup{mode=text,table-format=2.1,table-number-alignment=center}
\fontsize{9}{11}\selectfont
\renewcommand{\arraystretch}{1.12}
\begin{tabularx}{0.76\linewidth}{@{}>{\raggedright\arraybackslash}X S[table-column-width=64pt] S[table-column-width=64pt]@{}}
\toprule
\textbf{Method} & {TimeLens} & {LongVALE} \\
 & {mIoU (\%) $\uparrow$} & {mIoU (\%) $\uparrow$} \\
\midrule
\grp{3}{Proprietary Models} \\
Gemini-2.5-Pro & 58.2 & {--} \\
\addlinespace[4pt]
\grp{3}{Open-source Generalist Models} \\
Qwen2.5-VL-7B & 34.3 & {--} \\
Qwen3-VL-8B & 49.6 & {--} \\
Qwen3-VL-235B-A22B & {\underline{52.7}} & {--} \\
Qwen3.5-397B-A17B & \bfseries 53.5 & {--} \\
Qwen3-Omni-30B-A3B & {--} & 26.1 \\
\addlinespace[4pt]
\grp{3}{Open-source Agentic Models} \\
OmniAgent-7B & {--} & 39.1 \\
\addlinespace[2pt]
\rowcolor{OursRow}
\textbf{Open OmniAgent-7B} & 45.9 & {\underline{43.2}} \\
\rowcolor{OursRow}
\textbf{Open OmniAgent-30B} & 49.8 & \bfseries 47.5 \\
\bottomrule
\end{tabularx}
\end{table}

\begin{table}[!t]
\centering
\papertablestyle
\caption{Query-source comparison with 5K trajectories per source under the same pipeline. Accuracy (\%).}
\label{tab:query_source}
\setlength{\tabcolsep}{8pt}
\renewcommand{\arraystretch}{1.22}
\begin{tabular}{@{}lcccc@{}}
\toprule
\textbf{Query source} & Daily-Omni & LVOmniBench & OmniVideoBench & WorldSense \\
\midrule
OmniVideo-100K & 67.4 & 39.4 & 41.5 & 50.4 \\
\rowcolor{OursRow}
\textbf{Ours} & \textbf{68.4} & \textbf{40.3} & \textbf{42.2} & \textbf{51.8} \\
\bottomrule
\end{tabular}
\end{table}

\begin{table}[!t]
\centering
\papertablestyle
\caption{
Per-benchmark data scaling.
Models trained on OmniVideo-100K use direct SFT, while models trained on
Open OmniCoach use agentic SFT.
Avg.\ is the unweighted mean over the four benchmarks. 
Accuracy (\%) $\uparrow$.
}
\label{tab:data_scaling_full}
\fontsize{8}{10}\selectfont
\setlength{\tabcolsep}{3pt}
\renewcommand{\arraystretch}{1.14}
\begin{tabularx}{\linewidth}{
@{}>{\raggedright\arraybackslash}X
S[table-format=2.2]
S[table-format=2.2]
S[table-format=2.2]
S[table-format=2.2]
S[table-format=2.2]@{}
}
\toprule
\textbf{Training data}
& {WorldSense}
& {Daily-Omni}
& {OmniVideoBench}
& {LVOmniBench}
& {Avg.} \\
\midrule

\multicolumn{6}{@{}l}{\textit{OmniVideo-100K, direct SFT}} \\
\quad 25\%  & 54.98 & 73.02 & 45.10 & 42.00 & 53.78 \\
\quad 50\%  & 54.82 & 73.02 & 45.80 & 42.80 & 54.11 \\
\quad 75\%  & 54.70 & 75.10 & 44.80 & 43.29 & 54.47 \\
\quad 100\% & 52.84 & 72.51 & 46.40 & 42.50 & 53.56 \\

\midrule

\multicolumn{6}{@{}l}{\textit{Open OmniCoach, agentic SFT}} \\
\quad 25\%  & 52.60 & 69.30 & 42.70 & 42.41 & 51.75 \\
\quad 50\%  & 53.33 & 71.21 & 44.43 & 44.28 & 53.31 \\
\quad 75\%  & 54.06 & 73.11 & 46.17 & 46.10 & 54.86 \\
\quad 100\% & 54.79 & 75.02 & 47.03 & 47.21 & 56.01 \\

\bottomrule
\end{tabularx}
\end{table}

\textbf{Process-level evidence acquisition and error recovery.}
To evaluate intermediate evidence seeking and error recovery beyond final accuracy, we audit all 1,000 evaluation trajectories of Qwen3-Omni-30B-A3B variants on OmniVideoBench, which provides 3,577 human-annotated step-level evidence facts with timestamps and modalities (Table~\ref{tab:ovb_process_audit}).
From runtime tool logs, we measure the mean number of tool calls per query (\emph{Avg.\ Calls}), the share of malformed or duplicate requests returning no new media (\emph{Invalid / No-Prog.\ Calls}), the mean requested span duration (\emph{Avg.\ Span}), and the recall of annotated evidence timestamps within $\pm 2$\,s by spans of at most $30$\,s (\emph{Focused Hit $\le 30$\,s}).
In addition, given only the question, annotated facts, and per-turn JSON \texttt{observation} summaries defined in Appendix~\ref{app:tool_interface} (with model identity, reasoning, tool arguments, and final answers removed), a blind Gemini 3.6 Flash judge labels each $(\text{turn}, \text{fact})$ pair as \texttt{CORRECT}, \texttt{WRONG}, or \texttt{ABSENT}.
\emph{Error-to-Correct (+ Ans.\ Right)} measures, among trajectories where an early \texttt{observation} contradicts at least one annotated fact (\texttt{WRONG}), how often subsequent tool calls revise that fact to \texttt{CORRECT} in a later turn (and, in parentheses, also reach the correct final answer).
At comparable search depth to GSPO ($\lambda=0$), SEPO ($\lambda=0.1$) retrieves tighter spans ($42.1$\,s vs.\ $46.0$\,s) with higher focused recall ($62.9\%$ vs.\ $61.3\%$), makes fewer invalid or no-progress calls ($13.2\%$ vs.\ $15.1\%$), revises more early perceptual errors ($12.6\%$ vs.\ $12.0\%$), and more often converts those recoveries into correct final answers ($7.1\%$ vs.\ $5.0\%$).
Invalid or no-progress calls also decrease relative to GSPO on LVOmniBench, Video-MME-v2, WorldSense, and LVBench.

\textbf{RL checkpoint progression.}
Table~\ref{tab:rl_checkpoints} evaluates intermediate checkpoints at 50\%, 75\%, and 100\% of the RL training schedule starting from the Reflection SFT checkpoint in Table~\ref{tab:ablation} (which reports the peak checkpoint of each RL run).
On the four-benchmark average, SEPO outperforms GSPO at all three checkpoints ($+0.5$, $+0.6$, and $+0.8$ percentage points), with its 50\% checkpoint ($53.6\%$) already exceeding GSPO's best checkpoint ($53.3\%$).

\begin{table}[!t]
\centering
\papertablestyle
\caption{Process-level evidence acquisition and error recovery on OmniVideoBench (1,000 queries, Qwen3-Omni-30B-A3B).}
\label{tab:ovb_process_audit}
\fontsize{8}{10}\selectfont
\setlength{\tabcolsep}{3pt}
\renewcommand{\arraystretch}{1.14}
\begin{tabularx}{\linewidth}{@{}>{\raggedright\arraybackslash}Xccccc@{}}
\toprule
\textbf{Method} & \textbf{Avg.\ Calls} & \shortstack{\textbf{Invalid / No-Prog.}\\\textbf{Calls (\%) $\downarrow$}} & \shortstack{\textbf{Avg.}\\\textbf{Span (s) $\downarrow$}} & \shortstack{\textbf{Focused Hit}\\\textbf{$\le 30$\,s (\%) $\uparrow$}} & \shortstack{\textbf{Error-to-Correct}\\\textbf{(+ Ans.\ Right) (\%) $\uparrow$}} \\
\midrule
Reflection SFT & 3.9 & 8.4 & 43.3 & 58.2 & 9.0 (5.5) \\
+ GSPO ($\lambda=0$) & 5.4 & 15.1 & 46.0 & 61.3 & 12.0 (5.0) \\
\rowcolor{OursRow}
\textbf{+ SEPO ($\lambda=0.1$)} & 5.5 & \textbf{13.2} & \textbf{42.1} & \textbf{62.9} & \textbf{12.6 (7.1)} \\
\bottomrule
\end{tabularx}
\end{table}

\begin{table}[!t]
\centering
\papertablestyle
\caption{Evaluation across RL training checkpoints from the Reflection SFT checkpoint in Table~\ref{tab:ablation} (GSPO $\rightarrow$ SEPO, accuracy \%; higher value in each pair in bold).}
\label{tab:rl_checkpoints}
\fontsize{8}{10}\selectfont
\setlength{\tabcolsep}{3pt}
\renewcommand{\arraystretch}{1.14}
\begin{tabularx}{\linewidth}{@{}>{\raggedright\arraybackslash}Xccccc@{}}
\toprule
\textbf{RL Progress} & \textbf{Video-MME} & \textbf{Video-MME-v2} & \textbf{OmniVideoBench} & \textbf{LVOmniBench} & \textbf{Avg.} \\
\midrule
0\% (Reflection SFT) & 74.4 & 42.6 & 46.3 & 47.5 & 52.7 \\
50\% of schedule & 75.1 $\rightarrow$ \textbf{75.5} & 43.0 $\rightarrow$ \textbf{43.8} & 46.6 $\rightarrow$ \textbf{47.5} & \textbf{47.8} $\rightarrow$ 47.7 & 53.1 $\rightarrow$ \textbf{53.6} \\
75\% of schedule & 75.4 $\rightarrow$ \textbf{76.0} & 43.3 $\rightarrow$ \textbf{44.1} & 46.9 $\rightarrow$ \textbf{47.6} & 47.7 $\rightarrow$ \textbf{48.1} & 53.3 $\rightarrow$ \textbf{54.0} \\
100\% of schedule & 75.6 $\rightarrow$ \textbf{75.9} & 43.0 $\rightarrow$ \textbf{44.4} & 46.6 $\rightarrow$ \textbf{47.9} & 47.8 $\rightarrow$ \textbf{48.0} & 53.3 $\rightarrow$ \textbf{54.1} \\
\bottomrule
\end{tabularx}
\end{table}




\section{Task Synthesis Details}
\label{app:task_synthesis_details}

\subsection{Evidence Construction}
\label{app:evidence_construction}

An initial pass over coarse video chunks establishes a global entity table.
Audio and visual candidate boundaries are combined to divide the video into segments.
A VLM annotates speech, music, non-speech sounds, and visual content, allowing their temporal intervals to overlap.
Each observation records its interval, evidence type, description, and entity references.
The examiner revisits the original media with the draft annotations and produces a revised annotation.
Entity references are reconciled across segments, and unresolved identity assignments are removed.

Let $\mathcal E^{(t)}$ denote the corrected observations from segment $t$.
Starting from an empty timeline, we update the event representation sequentially:
\begin{equation}
\label{eq:timeline_update}
\mathcal H_t=\operatorname{Update}\bigl(\mathcal H_{t-1},\mathcal E^{(t)},\mathcal G\bigr).
\end{equation}
The update preserves earlier entries and may revise the latest entry when the next segment continues the same event.
New events are appended with references to their supporting observations.
Observations that do not create a new timeline entry remain available for later retrieval beyond the compressed event overview.

\subsection{Task Planning and Evidence Expansion}
\label{app:task_planning}

Task skills specify the required operation, candidate-selection rules, and supporting evidence.
For QA, the catalog contains 18 reasoning skills.
Candidate generation proceeds in two rounds, both using the complete timeline.
The second round additionally receives the first candidate bank and revises it to address omissions, incomplete candidates, and task mismatch.
For temporal grounding, candidate planning identifies audiovisual conditions suitable for localization, including conditions that recur across events.

Each candidate is associated with relevant events and potentially confounding events.
Expansion retrieves their explicitly referenced observations and eligible observations overlapping their temporal windows.
Evidence identifiers link the retrieved details to the underlying annotations.
Timeline summaries guide retrieval, while the detailed observations support task formulation.
Grounding additionally requires a full-video search after the query is fixed, because the planning timeline may compress repeated occurrences.

Task planning produces a task specification and references to candidate events.
The final instruction and answer are constructed after the relevant local evidence has been expanded.
Task-specific requirements guide both the selection of evidence and the form of the resulting task.

Each generated task is accompanied by its supporting evidence. Before accepting a sample, we ask a model to answer the final instruction from this evidence and compare its answer with the reference answer under task-specific criteria. The retained evidence intervals and modalities define $\mathcal A_x$.

\subsection{Question Answering}
\label{app:question_answering}

The question generator uses the selected evidence, relevant entities, and task requirements to formulate a draft question.
A separate solver answers the draft question from the corresponding original audio and video without annotation captions, a reference answer, or answer options.
Five distractors are then constructed to form a six-option question.

The draft undergoes a text-only audit for ambiguity, answer leakage, and inconsistent option construction, followed by a media-based revision using the selected and confounding events.
This revision may update the question, options, answer, and task label.
A fresh solver then answers the revised question without seeing the options.
A separate text-only assessment compares this answer with every option without access to the correct label.
The sample is accepted only when exactly one option matches and it agrees with the label.

The full-video check additionally verifies that omitted events do not change the reference answer.
Final samples retain source-segment references and their association with the original draft.

\subsection{Temporal Grounding}
\label{app:temporal_grounding}

For temporal grounding, planning identifies a candidate event or a recurring audiovisual relation.
We expand the relevant observations to establish the entities, audible and visible conditions, and potentially confounding events.
These details guide the formulation of a grounding query and the construction of its temporal answer.
Once the query has been specified, we search the complete video evidence for all matching occurrences.
Each occurrence is linked to supporting observations, from which its reference interval is derived.

For an occurrence supported by intervals $\{[s_i,e_i]\}$, its temporal label is constructed as their common intersection:
\begin{equation}
\label{eq:grounding_interval_intersection}
[s^*,e^*]=\left[\max_i s_i,\;\min_i e_i\right].
\end{equation}
Each occurrence requires both audio and visual evidence with a non-empty common intersection.
An independent ledger audit checks semantic consistency, evidence support, and completeness of the occurrence set.

Media verification then scans the full video in bounded batches.
A solver receives the query and media without reference intervals and predicts all matching occurrences.
Acceptance requires successful localization with both modalities and unsuccessful localization with either modality alone.
Localization is evaluated by matching predicted and reference occurrences one-to-one, requiring no unmatched occurrences and a temporal IoU above the specified threshold for every matched pair.

\subsection{Additional Task Formats}
\label{app:additional_task_formats}

During Stage-1 construction, the pipeline also generates three auxiliary text formats from the refined scripts.
For captioning, consecutive segments are grouped into coherent stories, and their detailed observations are used to generate descriptions.
Cloze tasks mask selected spans in these descriptions; blanks recoverable by a text-only solver are removed.
Timeline-description tasks organize the video into chronological event descriptions, with checks for temporal coverage, non-overlap, and support from the evidence annotations.
These timeline-description outputs are distinct from the internal event timeline used for task planning.

\section{Trajectory Synthesis Details}
\label{app:trajectory_synthesis_details}

\subsection{Tool Interface and Interaction Protocol}
\label{app:tool_interface}

Each trajectory starts with the task instruction, tool specifications, timestamped frames covering the video, and the full audio track when available.
Initial frames are sampled at 0.5 FPS.
If this produces more than 256 candidates, we uniformly subsample them to 256 frames.
Each initial frame uses a pixel budget of $224^2$.

At each step, the agent produces a JSON object containing \texttt{observation}, \texttt{think}, and \texttt{action}.
The observation describes the latest available media, while the reasoning combines relevant evidence and determines the next action.
The agent either invokes one tool or terminates with an \texttt{answer} action.
Direct answers are allowed at the first step, and the default interaction budget is 32 decision steps.
After tool use, observations include temporal references in \texttt{[MM:SS.m,MM:SS.m]} format.
Table~\ref{tab:tool_interface} summarizes the available tools and their per-call limits.

\begin{table}[!t]
\centering
\papertablestyle
\caption{Tool outputs and per-call limits.}
\label{tab:tool_interface}
\begin{tabularx}{\linewidth}{@{}>{\raggedright\arraybackslash}p{75pt}>{\centering\arraybackslash}p{37pt}>{\raggedright\arraybackslash}X>{\raggedright\arraybackslash}p{113pt}@{}}
\toprule
\textbf{Tool} & \textbf{Intervals} & \textbf{Returned evidence} & \textbf{Per-call limits} \\
\midrule
\texttt{get\_frames} & 1 & Sampled frames & At most 256 frames \\
\addlinespace[3pt]
\texttt{get\_audio} & 1 & Continuous audio & At most 256 seconds \\
\addlinespace[3pt]
\texttt{get\_clip} & 1 & Frames and audio from the same interval & At most 60 seconds \\
\addlinespace[3pt]
\texttt{get\_multi\_span} & \shortstack{2--8\\(ordered)} & Frames, audio, or both per interval & At most 256 frames and 128 seconds of audio in total \\
\bottomrule
\end{tabularx}
\end{table}

For visual retrieval, \texttt{level=coarse} and \texttt{level=detail} sample frames at 1 and 2 FPS, respectively.
Frames preserve their aspect ratio, with the longest side limited to 384 pixels by default or 1280 pixels under \texttt{resolution=high}.
High-resolution intervals are limited to 5 seconds, and a multi-span call may contain at most two such intervals.
Each multi-span entry also follows the limits of its corresponding single-interval tool.
Tool boundaries use \texttt{MM:SS.m} with one decimal place and must lie within the video; requested audio intervals must last at least 0.1 seconds.
For videos without audio, \texttt{get\_audio} is unavailable and \texttt{get\_clip} returns frames alone.

Multi-span results preserve the requested order and identify each interval as \texttt{span\_1}, \texttt{span\_2}, and so on.
The agent is instructed to describe the evidence or uncertainty for every returned interval and then compare them.
After a successful retrieval, subsequent model inputs retain raw media from the latest result while preserving earlier textual history.
All intervals returned by a multi-span call remain visible together.
Earlier media can be inspected again through a new tool call.

\subsection{Perceptual and Agentic Verification}
\label{app:trajectory_verification}

Perceptual verification audits newly introduced or revised claims, as well as claims repeated despite contradictory evidence.
Earlier observations provide context for identifying these changes; factual support comes from the media available at the current step.
Audio claims are checked against audio, visual claims against frames, and cross-modal associations against joint audio-visual evidence.
We distinguish unsupported claims from claims contradicted by the media.

A joint verifier receives the observation deltas and checks audio, visual, cross-modal, and agentic consistency against the media and trajectory state available at the current step.
For agentic verification, candidate findings identify a specific decision, the relevant evidence, and the suspected error.
The verifier confirms, rejects, or revises each candidate.
The criteria cover failure to revise contradicted hypotheses, repeated searches without new information, missing comparisons, inappropriate tool selection, and premature or delayed stopping.

\subsection{Diagnosis-Guided Refinement}
\label{app:trajectory_refinement}

At the selected step, we restore the original model-visible history and media.
The failed turn is supplied as an unexecuted draft; subsequent actions and observations from the original trajectory are excluded.
The teacher generates an updated observation, a concise draft preserving the original reasoning error, a correction, and a new action.
The draft and correction are combined into the reasoning field of the standard interaction format, and only the new action is executed.
Private diagnostic guidance is removed from the exported conversation.

The judge does not receive the reference answer during diagnosis or recovery.
A candidate is retained only when its final answer passes the same outcome check as the source trajectory.
Attempts restart from the same source state with different random seeds, stopping at the first accepted candidate or after max attempts.

Branching from failed student rollouts follows an online imitation-learning principle similar to DAgger~\citep{ross2011dagger}, supervising recovery on error states visited by the student policy.
Among the 17.4K retained reflection trajectories, 9.7K (spanning 3,926 queries) recover hard queries on which either all eight student rollouts failed or standard Stage-2 teacher sampling yielded no accepted trajectory.

\section{Evidence Reward Details}
\label{app:evidence_reward}

\subsection{Modality Matching and Task-Specific Scoring}
\label{app:evidence_scoring}

For task $x$, the score $q\in[0,1]$ compares a selected subset of retrieved evidence $\mathcal B$ with the task's evidence annotations $\mathcal A_x$.
The reward schema distinguishes visual, audio, and joint audio-visual evidence, denoted by $\mathrm V$, $\mathrm A$, and $\mathrm{AV}$.
A candidate's requested modality determines which evidence it can support; a joint audio-visual candidate may support visual, audio, or joint audio-visual targets when the target annotation permits that match.
Let $\mathcal M_x$ contain the annotated modalities with positive modality weights $\lambda_m$.
For modality $m\in\mathcal M_x$, write the target annotations as $\mathcal A_x^m=\{(\tau_j,w_j)\}_{j=1}^{J_m}$, where $\tau_j$ is a temporal interval, $w_j>0$ its importance weight, and $J_m$ the number of targets.
Let $P_m$ be the union of intervals in the selected subset $\mathcal B$ with modality $m$.
The combined score is
\begin{equation}
\label{eq:modality_weighted_score}
q(\mathcal B,\mathcal A_x)
=\frac{\sum_{m\in\mathcal M_x}\lambda_m q_m(P_m,\mathcal A_x^m)}
{\sum_{m\in\mathcal M_x}\lambda_m},
\end{equation}
where $q_m$ is the task-specific score below.
An annotated modality with no selected candidate receives zero; modalities absent from the targets do not contribute.
The scorer defaults to equal modality weights and unit target weights.
Overlapping intervals are merged before computing support duration, denoted by $|\cdot|$.

\paragraph{Question answering.}
For a fixed modality, define the target union $G_m=\bigcup_{j=1}^{J_m}\tau_j$, total weight $Z_m=\sum_{j=1}^{J_m}w_j$, and weighted support duration $D_m=\sum_{j=1}^{J_m}w_j|\tau_j|$.
Coverage $c_m$ and locality $\ell_m$ are
\begin{equation}
\label{eq:qa_coverage_locality}
\begin{aligned}
c_m&=\frac{1}{Z_m}\sum_{j=1}^{J_m}w_j
\frac{|P_m\cap\tau_j|}{|\tau_j|},\\
\ell_m&=\begin{cases}
\left(1+\dfrac{|P_m\setminus G_m|}{D_m}\right)^{-1},
& |P_m\cap G_m|>0,\\
0,&\text{otherwise}.
\end{cases}
\end{aligned}
\end{equation}
The QA score is $q_m=\alpha c_m+(1-\alpha)\ell_m$.
Coverage rewards retrieval of annotated support, while locality favors a more focused interval once support has been reached.
The scorer uses $\alpha=0.5$ by default.

\paragraph{Temporal grounding.}
TG targets specify the interval set to be localized.
For non-empty temporal interval unions $P$ and $G$, let $\phi(P,G)$ denote the grounding reward for valid predictions in TimeLens2~\citep{zhu2026timelens2}, scaled by $1/2$ to lie in $[0,1]$.
For equal target weights, $q_m=\phi(P_m,G_m)$.
For unequal weights, the scorer uses $q_m=Z_m^{-1}\sum_{j=1}^{J_m}w_j\phi(P_m,\tau_j)$.
An empty retrieved support receives zero.

\subsection{Selected Evidence State and Incremental Credit}
\label{app:evidence_state}

For task $x$, let $\mathcal C_t$ denote the collection of retrieved intervals and their modalities available after step $t$.
The ideal selected-evidence utility is
\begin{equation}
\label{eq:optimal_evidence_utility}
U_t^\star=\max_{\mathcal B\subseteq\mathcal C_t}q(\mathcal B,\mathcal A_x).
\end{equation}
The evidence scorer starts with $\mathcal C_0=\emptyset$ and $U_0=0$, accumulating the temporal intervals and modalities returned by successful tool calls and retaining this metadata across turns even after earlier raw media leave the active context.
The initial overview is not added to $\mathcal C_t$: because all $G=8$ rollouts of a query share the same coarse initial overview (Appendix~\ref{app:tool_interface}), the scorer evaluates only the incremental local evidence retrieved by tool calls.
On OmniVideoBench (3,577 annotated facts across 1,000 trajectories), initial Turn-0 \texttt{observation} summaries capture 15.1\% of required facts, whereas subsequent tool calls add 22.5 percentage points of newly grounded facts (37.6\% cumulative recall), confirming that retrieved spans supply task-relevant information beyond the initial overview.

The subset maximization separates by modality because $q$ is a weighted sum of independent modality scores.
The implementation enumerates all subsets when a modality has at most 12 candidates, preferring fewer candidates when scores tie.
Above this limit, it ranks candidates by their singleton score, retains the top 12, and enumerates subsets of this reduced collection.
Equal singleton scores are ordered by candidate identifier.

Let $\widetilde U_t$ be the utility returned by this bounded search at step $t$.
Because the reduced candidate set can change, the implemented state retains the previous utility whenever a new approximation decreases it:
\begin{equation}
\label{eq:monotone_evidence_utility}
U_0=\widetilde U_0,\qquad
U_t=\max\{U_{t-1},\widetilde U_t\},\qquad
\Delta U_t=U_t-U_{t-1}.
\end{equation}
For trajectory $i$ and turn $k$, the corresponding quantities are written as $U_{i,k}$ and $\Delta U_{i,k}$ in Sec.~\ref{subsec:evidence_guided_rl}.

\subsection{Step-Level Rewards and Policy Optimization}
\label{app:step_reward_optimization}

\paragraph{Trajectory reward.}
The answer reward $R_i^{\mathrm{ans}}$ depends on the task type.
Multiple-choice answers receive 1 for a matching option label and 0 otherwise.
Free-form answers first undergo normalized exact matching; unmatched answers are judged against the question and reference answer, receiving 1 if correct and 0 otherwise.
For temporal grounding, the answer reward is $\phi(P,G)$ defined in Appendix~\ref{app:evidence_scoring}, where $P$ and $G$ are the predicted and reference interval unions.
An absent or invalid final answer receives zero answer reward.
The protocol reward $R_i^{\mathrm{fmt}}$ is $+0.1$ when the final answer uses the canonical format and the trajectory contains no invalid or no-progress actions, and $-0.1$ otherwise.
We sample $G=8$ trajectories per instruction and subtract their mean total reward to obtain $A_i$, without dividing by the group standard deviation.
\begin{equation}
\label{eq:trajectory_advantage}
R_i=R_i^{\mathrm{ans}}+R_i^{\mathrm{fmt}},
\qquad
A_i=R_i-\frac{1}{G}\sum_{j=1}^{G}R_j.
\end{equation}

\paragraph{Process score.}
Extending the evidence-gain formulation in Sec.~\ref{subsec:evidence_guided_rl} with turn-level protocol penalties, each generated turn $k$ receives a process score $p_{i,k}$:
\begin{equation}
\label{eq:process_score}
\begin{aligned}
p_{i,k}&=0.2\,\Delta U_{i,k}
-0.1\,I_{i,k}^{\mathrm{invalid}}
-0.05\,I_{i,k}^{\mathrm{repeat}},\\
z_{i,k}&=\frac{p_{i,k}-\mu_i}{\sigma_i+\delta}.
\end{aligned}
\end{equation}
Here, $I_{i,k}^{\mathrm{invalid}}$ indicates a malformed tool call or out-of-range timestamp, and $I_{i,k}^{\mathrm{repeat}}$ indicates an exact duplicate request matching the interval, modality, and sampling settings of an earlier call; a request with a different modality or sampling precision is not marked as a repeat solely because its interval overlaps an earlier call.
The statistics $\mu_i$ and $\sigma_i$ are the mean and population standard deviation of $p_{i,k}$ across generated turns in trajectory $i$, with $\delta=10^{-6}$ and $\lambda=0.1$.
For tasks without evidence annotations, $\Delta U_{i,k}=0$ while the protocol indicators remain active.
The process score only rescales turn advantages and is neither added to $R_i$ nor used for trajectory filtering.

\paragraph{Within-trajectory adjustment.}
Let $K_i$ be the number of generated turns in trajectory $i$, including the final-answer turn.
The normalization statistics are
\begin{equation}
\label{eq:process_score_statistics}
\mu_i=\frac{1}{K_i}\sum_{k=1}^{K_i}p_{i,k},
\qquad
\sigma_i=\sqrt{\frac{1}{K_i}\sum_{k=1}^{K_i}(p_{i,k}-\mu_i)^2}.
\end{equation}
Replacing the standardized step gain in Sec.~\ref{subsec:evidence_guided_rl} with $z_{i,k}$ gives $\widetilde A_{i,k}=A_i+\lambda |A_i|\operatorname{clip}(z_{i,k},-1,1)=A_i w_{i,k}$, where
\begin{equation}
\label{eq:turn_advantage_weight}
w_{i,k}=1+\lambda\operatorname{clip}\!\left(\operatorname{sign}(A_i)z_{i,k},-1,1\right)
\in[0.9,1.1].
\end{equation}
When the agent answers directly from the initial overview in a single turn (or when all turns share the same process score), $w_{i,k}=1$.
When $A_i=0$, $\widetilde A_{i,k}=0$; because $w_{i,k}>0$, the adjustment preserves the sign of $A_i$.

\paragraph{Policy update.}
Each rollout batch performs a single on-policy gradient update (one PPO epoch).
In the trainer, each generated turn is passed as a response segment whose tokens inherit $\widetilde A_{i,k}$, and response-token losses are summed and normalized by the total number of trainable response tokens $N_i$ in trajectory $i$.
Under a single on-policy update per rollout batch, the importance ratio equals $1$ when evaluating the gradient, so this turn-segmented implementation yields the exact same token-level policy gradient as the GSPO-token objective in Sec.~\ref{subsec:evidence_guided_rl}, with ratio clipping bounds $[0.8, 1.28]$ acting as numerical safeguards.
We include a KL penalty against the frozen SFT reference policy with coefficient $0.001$.
Invalid and no-progress action tokens remain trainable, whereas operational failures are excluded from the policy loss.
The GSPO baseline shares the identical codebase, trajectory rewards, protocol checks, and KL coefficient, setting $\lambda=0$ so that $w_{i,k}=1$ for all turns.

\section{Dataset Composition}
\label{app:dataset_composition}

\begin{figure}[!htb]
\centering
\includegraphics[width=\linewidth]{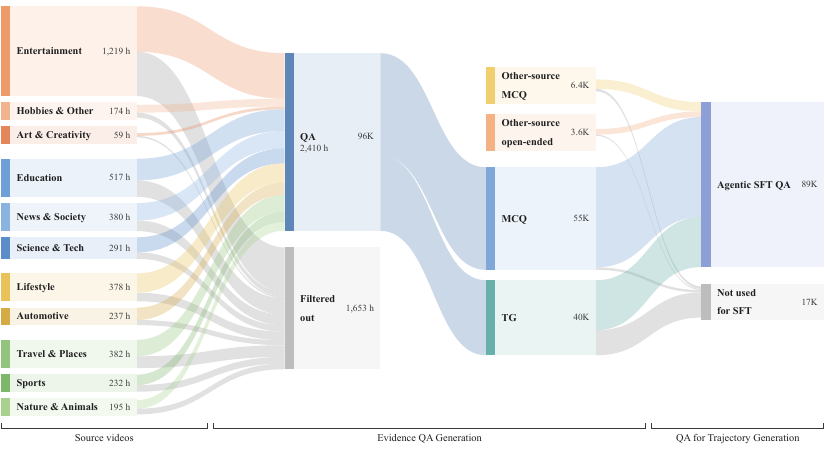}
\caption{
\textbf{QA data flow behind OmniCoach-240K.}
Band widths encode video hours between the first two columns and QA counts afterward.
The brackets mark the source videos, Stage-1 (Hierarchical Evidence QA Generation), and the QA samples used for Stage-2 and Stage-3 trajectory generation; other-source QA samples join in the third column.
\emph{Filtered out}: source videos without any retained QA sample.
\emph{Agentic SFT QA}: QA samples covered by at least one retained Stage-2 teacher or Stage-3 reflection trajectory.
}
\label{fig:data_flow_sankey}
\end{figure}

Fig.~\ref{fig:data_flow_sankey} traces the QA samples behind OmniCoach-240K, from source videos to agentic SFT.
The source videos span 11 content domains.
Stage-1 retains 95.7K QA samples (55.3K MCQ and 40.4K TG) from videos totaling 2,410 hours; the remaining 1,653 hours of video yield no retained QA sample.
These QA samples are used for trajectory generation in Stage-2 and Stage-3, together with 10K other-source public QA samples (6.4K MCQ and 3.6K open-ended QA, i.e., the free-form (FF) format in Table~\ref{tab:dataset_comparison}).
Of the 105.7K QA samples used for trajectory generation, 88.9K are covered by at least one retained teacher or reflection trajectory and are used for agentic SFT; the remaining 16.8K have no retained trajectory and are not used for SFT.

\section{Limitations and Future Work} 
Our study is limited to a few omni-video backbones and benchmarks, while trajectory construction relies on strong teacher models and multi-stage verification, incurring additional cost.
Reflection trajectories improve error recovery but do not cover all possible failure modes.
Future work will explore (1) more efficient data construction within the data model in the loop, (2) stronger self-improvement mechanisms for omni-video agents, including enhance our stage 3 design in the data learning agent loop.

\end{document}